\documentclass[11pt]{article}

\usepackage{eacl2014}
\usepackage{times}
\usepackage{hhline}
\usepackage{latexsym}
\usepackage{multirow}
\usepackage{url}	
\usepackage{graphicx}
\usepackage{amssymb}

\title{Learning Dictionaries for Named Entity Recognition using \\ Minimal Supervision}

\author{Arvind Neelakantan \\
  Department of Computer Science \\
University of Massachusetts, Amherst \\ Amherst, MA, 01003 \\
  {\tt arvind@cs.umass.edu} \\\And
 Michael Collins \\
  Department of Computer Science \\ Columbia University \\ New-York, NY 10027, USA \\
  {\tt mcollins@cs.columbia.edu} \\}

\date{}

\begin{document}

\maketitle
\begin{abstract}
This paper describes an approach for automatic construction of dictionaries for Named Entity Recognition (NER) using large amounts of unlabeled data and a few seed examples.  
We use Canonical Correlation Analysis (CCA)  to obtain lower dimensional embeddings (representations) for \emph{candidate phrases} and classify these phrases using a small number of labeled examples. Our method achieves 16.5\% and 11.3\%  F-1 score improvement over co-training on disease and virus NER respectively.  We also show that by adding candidate phrase embeddings  as features in a sequence tagger gives better performance compared to using word embeddings.
\end{abstract}

\section{Introduction}
Several works (e.g., Ratinov and Roth, 2009; Cohen and Sarawagi, 2004) have shown that injecting dictionary matches as features in a sequence tagger results in significant gains in NER performance. However, building these dictionaries requires a huge amount of human effort and it is often difficult to get good coverage  for many named entity types. The problem is more severe when we consider named entity types such as gene, virus and disease, because of the large (and growing) number of names in use, the fact that the names are heavily abbreviated and multiple names are used to refer to the same entity \cite{leaman,ncbi}. Also, these dictionaries can only be built by domain experts, making the process very expensive.

This paper describes an approach for automatic construction of dictionaries for NER using large amounts of unlabeled data and a small number of seed examples.  Our approach consists of two steps. First, we collect a high recall, low precision list of \emph{candidate phrases} from the large unlabeled  data collection for every named entity type using simple rules. In the second step, we construct an accurate dictionary of named entities by removing the noisy candidates from the list obtained in the first step. This is done by learning a classifier using the lower dimensional, real-valued CCA \cite{cca} embeddings of the candidate phrases as features and training it using a small number of labeled examples. The classifier we use is a binary SVM which predicts whether a candidate phrase is a named entity or not.

We compare our method to a widely used semi-supervised algorithm based on co-training \cite{blum}.  The dictionaries are first evaluated on virus \cite{genia} and disease \cite{ncbi} NER by using them directly in dictionary based taggers. We also give results comparing the dictionaries produced by the two semi-supervised approaches with dictionaries that are compiled manually. The effectiveness of the dictionaries are also measured by injecting dictionary matches as features in a Conditional Random Field (CRF) based tagger. The results indicate that our approach with minimal supervision produces dictionaries that are comparable to  dictionaries compiled manually. Finally, we also compare the quality of the candidate phrase embeddings with word embeddings \cite{dhillon_lrmv} by adding them as features in a CRF based sequence tagger.      

\section{Background}
We first give background on Canonical Correlation Analysis (CCA), and then give background on CRFs for the NER problem.
\subsection{Canonical Correlation Analysis (CCA)}
The input to CCA consists of $n$ paired observations  $(x_{1}, z_{1}),\ldots,(x_{n}, z_{n})$ where $x_{i}  \in \mathbb{R} ^{d_{1}} , z_{i} \in \mathbb{R} ^{d_{2}}  \;  (\forall  i \in \{1,2,\ldots,n\})$ are the feature representations for the two views of a data point. CCA simultaneously learns projection matrices $\Phi_{1} \in \mathbb{R}^{d_{1} \times k} , \Phi_{2} \in \mathbb{R}^{d_{2} \times k}$ ($k$ is a small number) which are used to obtain the lower dimensional representations  $(\bar{x}_{1}, \bar{z}_{1}),\ldots,(\bar{x}_{n}, \bar{z}_{n})$ where $\bar{x}_{i}=\Phi_{1}^{\mathsf{T}}x_{i}  \in \mathbb{R} ^{k} , \bar{z}_{i}=\Phi_{2}^{\mathsf{T}}z_{i} \in \mathbb{R} ^{k}, \;  \forall  i \in \{1,2,\ldots,n\}$. $\Phi_{1}, \Phi_{2}$ are chosen to maximize the correlation between $\bar{x}_{i}$ and $\bar{z}_{i},  \;  \forall  i \in \{1,2,\ldots,n\}$.

Consider the setting where we have a label for the data point along with it's two views and either view is sufficient to make accurate predictions. \newcite{sham} and \newcite{convex_loss} give strong theoretical guarantees when the lower dimensional embeddings from CCA are used for predicting the label of the data point. This setting is similar to the one considered in co-training \cite{collins_co_training} but there is no assumption of independence between the two views of the data point. Also, it is an exact algorithm unlike the algorithm given in \newcite{collins_co_training}. Since we are using lower dimensional embeddings of the data point for prediction, we can learn a predictor with fewer labeled examples.

\subsection{CRFs for Named Entity Recognition}

CRF based sequence taggers have been used for a number of NER tasks (e.g., McCallum and Li, 2003) and in particular for biomedical NER (e.g., McDonald and Pereira, 2005; Burr Settles, 2004)  because they allow a great deal of flexibility in the features which can be included. The input to a CRF tagger is a sentence ($w_{1}, w_{2}, \ldots,w_{n}$) where $w_{i}, \; \forall i \in \{1,2,\ldots,n\}$ are words in the sentence. The output is a sequence of tags $y_{1}, y_{2}, \ldots, y_{n}$ where $y_{i} \in \{\texttt{B, I, O}\}, \; \forall i \in \{1,2,\ldots,n\}$. $\texttt{B}$ is the tag given  to the first word in a named entity,  $\texttt{I}$ is the tag given to all words except the first word in a named entity  and $\texttt{O}$ is the tag given to all other words. We used the standard NER baseline features (e.g., Dhillon et al., 2011; Ratinov and Roth, 2009) which include:
\begin{itemize}
\item{Current Word $w_{i}$ and its lexical features which include whether the word is capitalized and whether all the characters are capitalized. Prefix and suffixes of the word $w_{i}$ were also added.}
\item{Word tokens in window of size two around the current word which include $w_{i-2}, w_{i-1},  w_{i+1}, w_{i+2}$ and also the capitalization pattern in the window.}
\item{Previous two predictions $y_{i-1}$ and $y_{i-2}$.}
\end{itemize}
The effectiveness of the dictionaries are evaluated by adding dictionary matches as features along with the baseline features \cite{dan_roth,cohen} in the CRF tagger. We also compared the quality of the candidate phrase embeddings with the word-level embeddings by adding them as features \cite{dhillon_lrmv} along with the baseline features in the CRF tagger. 

\section{Method}
This section describes the two steps in our approach: obtaining candidate phrases and classifying them.
\subsection{Obtaining Candidate Phrases} 
We used the full text of 110,369 biomedical publications in the BioMed Central corpus\footnote{The corpus can be downloaded at http://www.biomedcentral.com/about/datamining} to get the high recall, low precision  list of candidate phrases. The advantages  of using this huge collection of publications are obvious: almost all (including rare) named entities related to the biomedical domain will be mentioned and contains more recent developments than a structured resource like Wikipedia. The challenge however is that these publications are unstructured and hence it is a difficult task to construct accurate dictionaries using them with minimal supervision.
 
The list of virus candidate phrases were obtained by extracting phrases that occur between ``the'' and ``virus'' in  the simple pattern ``the ... virus'' during a single pass over the unlabeled document collection. This noisy list  had a lot of virus names such as \emph{influenza}, \emph{human immunodeficiency} and \emph{Epstein-Barr} along with phrases that are not virus names, like \emph{mutant}, \emph{same}, \emph{new}, and so on.

 A similar rule like ``the ... disease'' did not give a good coverage of disease names since it is not the common way of how diseases are mentioned in publications. So we took a different approach to obtain the noisy list of disease names.  We collected every sentence in the unlabeled data collection that has the word ``disease'' in it and extracted  noun phrases\footnote{Noun phrases were obtained using http://www.umiacs.umd.edu/~hal/TagChunk/}  following the patterns ``diseases like ....'', ``diseases such as ....'' , ``diseases including ....'' , ``diagnosed with ....'', ``patients with ....'' and ``suffering from ....''.

\subsection{Classification of Candidate Phrases}
Having found the list of candidate phrases, we now describe how noisy words are filtered out from them.  We gather (\emph{spelling}, \emph{context}) pairs for every instance of a candidate phrase in the unlabeled data collection. \emph{spelling} refers to the candidate phrase itself while \emph{context} includes three words each to the left and the right of the candidate
phrase in the sentence. The \emph{spelling} and the  \emph{context} of the candidate phrase provide a natural split into two views which multi-view algorithms like co-training and CCA can exploit. The only supervision in our method is to provide a few \emph{spelling} seed examples (10 in the case of virus, 18 in the case of disease), for example, \emph{human immunodeficiency} is a virus and \emph{mutant} is not a virus. 

\subsubsection{Approach using CCA embeddings}
We use CCA described in the previous section to obtain lower dimensional embeddings for the candidate phrases using the (\emph{spelling}, \emph{context}) views. Unlike previous works such as \newcite{dhillon_lrmv} and \newcite{dhillon_two}, we use CCA to learn embeddings for candidate phrases instead of all words in the vocabulary so that we don't miss named entities which have two or more words.  

Let the number of (\emph{spelling}, \emph{context}) pairs be $n$ (sum of total number of instances of every candidate phrase in the unlabeled data collection).  First, we map the \emph{spelling} and \emph{context} to high-dimensional feature vectors. For the \emph{spelling} view, we define a feature for every candidate phrase and also a boolean feature which indicates whether the phrase is capitalized or not. For the \emph{context} view, we use features similar to \newcite{dhillon_lrmv} where a feature for every word in the \emph{context}  in conjunction with its position is defined. Each of the $n$ (\emph{spelling}, \emph{context}) pairs are mapped to a pair of high-dimensional feature vectors to get $n$ paired observations $(x_{1}, z_{1}),\ldots,(x_{n}, z_{n})$ with $x_{i}  \in \mathbb{R} ^{d_{1}} , z_{i} \in \mathbb{R} ^{d_{2}}, \;  \forall  i \in \{1,2,\ldots,n\}$ ($d_{1}, d_{2}$ are the feature space dimensions of the \emph{spelling} and \emph{context} view respectively). Using CCA\footnote{Similar to \newcite{dhillon_two} we used the method given in \newcite{approx} to perform the SVD computation in CCA  for practical considerations.}, we learn the projection matrices $\Phi_{1} \in \mathbb{R}^{d_{1} \times k} , \Phi_{2} \in \mathbb{R}^{d_{2} \times k}$ ($k << d_{1}$ and $k << d_{2}$ )  and obtain \emph{spelling} view projections $\bar{x}_{i}=\Phi_{1}^{\mathsf{T}}x_{i}  \in \mathbb{R} ^{k}, \forall  i \in \{1,2,\ldots,n\} $. The k-dimensional \emph{spelling} view projection of any instance of a candidate phrase is used as it's embedding\footnote{Note that a candidate phrase gets the same \emph{spelling} view projection across it's different instances since the \emph{spelling} features of a candidate phrase are identical across it's instances.}. 

The k-dimensional candidate phrase embeddings are used as features to learn a binary SVM with the seed \emph{spelling} examples given in figure 1 as training data. The binary SVM predicts whether a candidate phrase is a named entity or not. Since the value of k is small, a small number of labeled examples are sufficient to train an accurate classifier. The learned SVM is used to filter out the noisy phrases from  the list of  candidate phrases obtained in the previous step. 

To summarize, our approach for classifying candidate phrases has the following steps:
\begin{itemize}
\item{Input: $n$ (\emph{spelling}, \emph{context}) pairs, \emph{spelling} seed examples.  }
\item{Each of the $n$ (\emph{spelling}, \emph{context}) pairs are mapped to a pair of high-dimensional feature vectors to get $n$ paired observations $(x_{1}, z_{1}),\ldots,(x_{n}, z_{n})$ with $x_{i}  \in \mathbb{R} ^{d_{1}} , z_{i} \in \mathbb{R} ^{d_{2}}, \;  \forall  i \in \{1,2,\ldots,n\}$.}
\item{Using CCA, we learn the projection matrices $\Phi_{1} \in \mathbb{R}^{d_{1} \times k} , \Phi_{2} \in \mathbb{R}^{d_{2} \times k}$ and obtain \emph{spelling} view projections $\bar{x}_{i}=\Phi_{1}^{\mathsf{T}}x_{i}  \in \mathbb{R} ^{k}, \forall  i \in \{1,2,\ldots,n\} $.}
\item{The embedding of a candidate phrase is given by the  k-dimensional \emph{spelling} view projection of any instance of the candidate phrase.}
\item{We learn a binary SVM with the candidate phrase embeddings as features  and the \emph{spelling} seed examples given in figure 1 as training data. Using this SVM, we predict whether a candidate phrase is a named entity or not. }
 \end{itemize}
\begin{figure*}	
\centering	
    \fbox{\includegraphics[trim= 85 610 75 70,clip,width=0.95\textwidth]{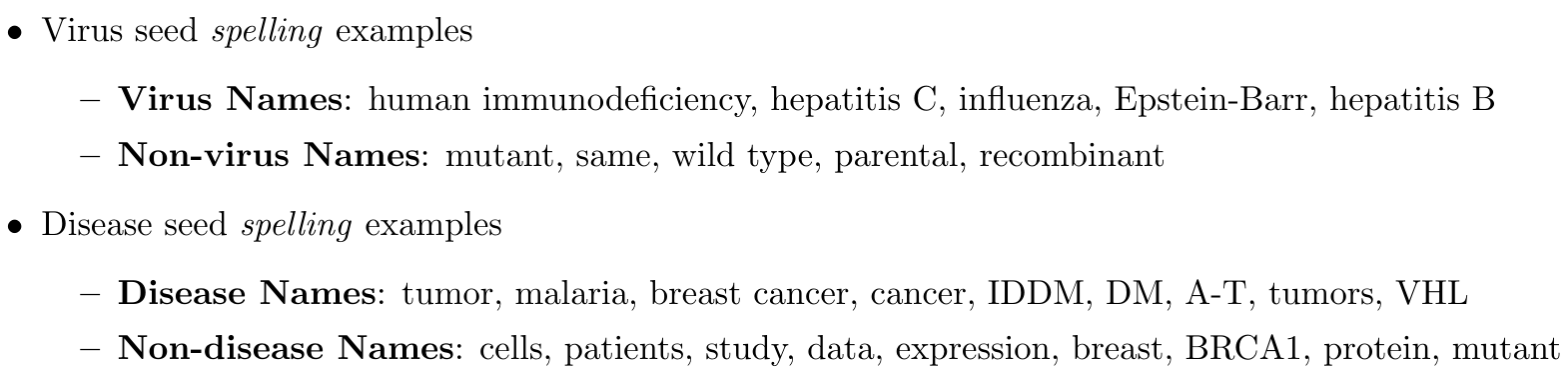}}
     \caption{Seed \emph{spelling} examples}
\end{figure*}

\subsubsection{Approach based on Co-training}
We discuss here briefly the DL-CoTrain algorithm \cite{collins_co_training} which is based on co-training \cite{blum}, to classify candidate phrases. We compare our approach using CCA embeddings with this approach. Here, two decision list of rules are learned simultaneously one using the \emph{spelling} view and the other using the \emph{context} view.  The rules using the 
\emph{spelling} view are of the form: full-string=human immunodeficiency $\rightarrow$ Virus, full-string=mutant $\rightarrow$ Not\_a\_virus and so on. In the \emph{context} view, we used bigram\footnote{We tried using unigram rules but they were very weak predictors and the performance of the algorithm was poor when they were considered.} rules where we considered all possible bigrams using the \emph{context}. The rules are of two types: one which gives a positive label, for example, full-string=human immunodeficiency $\rightarrow$ Virus and the other which gives a negative label, for example,  full-string=mutant $\rightarrow$ Not\_a\_virus. The DL-CoTrain algorithm is as follows:	 
\begin{itemize}
\item{Input: (\emph{spelling}, \emph{context}) pairs for every instance of a candidate phrase in the corpus, $m$ specifying the number of rules to be added in every iteration, precision threshold $\epsilon$, \emph{spelling} seed examples.}
\item{Algorithm:
	\begin{description}
	\item{ 1. Initialize the \emph{spelling} decision list using the \emph{spelling} seed examples given in figure 1 and set $i=1$.}
	\item{ 2. Label the entire input collection using the learned decision list of \emph{spelling} rules.}
	\item{ 3. Add $i \times m$ new \emph{context} rules of each type to the decision list of \emph{context} rules using the current labeled data. The rules are added using the same criterion as given in \newcite{collins_co_training}, i.e., among the rules whose strength is greater than the precision threshold $\epsilon$, the ones which are seen more often with the corresponding label in the input data collection are added.  }
	\item{ 4. Label the entire input collection using the learned decision list of \emph{context} rules.}
	\item{ 5. Add $i \times m$ new spelling rules of each type to the decision list of \emph{spelling} rules using the current labeled data. The rules are added using the same criterion as in step 3. Set $i=i+1$. If rules were added in the previous iteration, return to step 2.}
	\end{description}
}
\end{itemize}
The algorithm is run until no new rules are left to be added. The \emph{spelling} decision list along with its strength \cite{collins_co_training} is used to construct the dictionaries. The phrases present in the \emph{spelling} rules which give a positive label and whose strength is greater than the precision threshold,  were added to the dictionary of named entities. We found the parameters m and $\epsilon$ difficult to tune and they could significantly affect the performance of the algorithm. We give more details regarding this in the  experiments section.

\section{Related Work}
 Previously, \newcite{collins_co_training} introduced a multi-view, semi-supervised algorithm based on co-training \cite{blum} for collecting names of people, organizations and locations. This algorithm makes a strong independence assumption about the data and employs many heuristics to greedily optimize an objective function. This greedy approach also introduces new parameters that are often difficult to tune. 

In other works such as  \newcite{wiki_2} and \newcite{wiki_1} external structured resources like Wikipedia have been used to construct dictionaries. Even though these methods are fairly successful they suffer from a number of drawbacks especially in the biomedical domain. The main drawback of these approaches is that it is very difficult to accurately disambiguate ambiguous entities especially when the entities are abbreviations \cite{wiki_1}. For example, \emph{DM} is the abbreviation for the disease \emph{Diabetes Mellitus} and the disambiguation page for \emph{DM} in Wikipedia associates it to more than 50 categories since \emph{DM} can be expanded to \emph{Doctor of Management}, \emph{Dichroic mirror}, and so on, each of it belonging to a different category. Due to the rapid growth of Wikipedia,  the number of entities that have disambiguation pages is growing fast and it is increasingly difficult to retrieve the article we want. Also, it is tough to understand these approaches from a theoretical standpoint.

\newcite{dhillon_lrmv}  used CCA to learn word embeddings and added them as features in a sequence tagger. They show that CCA learns better word embeddings than  CW embeddings \cite{cw}, Hierarchical log-linear (HLBL) embeddings \cite{hlbl} and embeddings learned from many other techniques for NER and chunking. Unlike PCA, a widely used dimensionality reduction technique, CCA is invariant to linear transformations of the data.  Our approach is motivated by the theoretical result  in \newcite{sham} which is developed in the co-training setting. We directly use the CCA embeddings to predict the label of a data point instead of using them as features in a sequence tagger. Also, we learn CCA embeddings for candidate phrases instead of all words in the vocabulary since named entities often contain more than one word. \newcite{dhillon_two} learn a multi-class SVM using the CCA word embeddings to predict the POS tag of a word type. We extend this technique to NER by learning a binary SVM using the CCA embeddings of a high recall, low precision list of candidate phrases to predict whether a candidate phrase is a named entity or not.

\section{Experiments}
In this section, we give experimental results on virus and disease NER. 
\subsection{Data}
The noisy lists of both virus and disease names were obtained from the BioMed Central corpus. This corpus was also used to get the collection of (\emph{spelling}, \emph{context}) pairs which are the  input to the CCA procedure and the DL-CoTrain algorithm described in the previous section. We  obtained CCA embeddings for the $100,000$ most frequently occurring word types in this collection along with  every word type present in the training and development data of the virus and the disease NER dataset. These word embeddings are similar to the ones described in \newcite{dhillon_lrmv} and \newcite{dhillon_two}.

We used the virus annotations in the GENIA corpus \cite{genia} for our experiments. The dataset contains 18,546 annotated sentences. We randomly selected 8,546 sentences for training and the remaining sentences were randomly split equally into development and testing sentences. The training sentences  are used only for experiments with the sequence taggers. Previously,  \newcite{virus} tested their HMM-based named entity recognizer on this data. For disease NER, we used the recent disease corpus \cite{ncbi} and used the same training, development and test data split given by them. We used a sentence segmenter\footnote{https://pypi.python.org/pypi/text-sentence/0.13} to get sentence segmented data and Stanford Tokenizer\footnote{http://nlp.stanford.edu/software/tokenizer.shtml} to tokenize the data.  Similar to \newcite{ncbi}, all the different disease categories were flattened into one single category of disease mentions. The development data was used to tune the hyperparameters and the methods were evaluated on the test data.

\subsection{ Results using a dictionary-based tagger}

\begin{table*} [t]
\centering
\begin{tabular}{ | c | c | c | c | c | c | c |}
\hline
{\multirow{2}{*}{Method}} & \multicolumn{3}{c|} {Virus NER} & \multicolumn{3}{c|} {Disease NER} \\
\hhline{~------}
& Precision & Recall & F-1 Score & Precision & Recall & F-1 Score \\
\hline
Candidate List & 2.20 & {\bf 69.58} & 4.27 & 4.86 & {\bf 60.32} & 8.99 \\
\hline
Manual & 42.69 & 68.75 & 52.67 & 51.39 & 45.08 & {\bf 48.03}\\  
\hline
Co-Training & 48.33 & 66.46 & 55.96 & {\bf 58.87} & 23.17 & 33.26\\
\hline
CCA & {\bf 57.24} & 68.33 & {\bf 62.30} & 38.34 & 44.55 & 41.21 \\
\hline
\end{tabular}
\caption{Precision, recall, F- 1 scores of dictionary-based taggers }
\end{table*}
First, we compare the dictionaries compiled using different methods by using them  directly in a dictionary-based tagger. This is a simple and informative way to understand the quality of the dictionaries before using them in a CRF-tagger. Since these taggers can be trained using a handful of training examples, we can use them to build NER systems even when there are no labeled sentences to train. The input to a dictionary tagger is a list of named entities and a sentence. If there is an exact match between a phrase in the input list to the words in the given sentence then it is tagged as a named entity. All other words are labeled as non-entities. We evaluated the performance of the following methods for building dictionaries:

\begin{itemize}
\item{{\bf Candidate List}: This dictionary contains all the candidate phrases that were obtained using the method described in  Section 3.1. The noisy list of virus candidates and disease candidates had 3,100 and 60,080 entries respectively.  }
\item{{\bf Manual}: Manually constructed dictionaries, which requires a large amount of human effort, are employed for the task. We used the list of virus names given in Wikipedia\footnote{http://en.wikipedia.org/wiki/List\_of\_viruses}. Unfortunately, abbreviations of virus names are not present in this list and we could not find any other more complete list of virus names. Hence, we constructed abbreviations by concatenating the first letters of all the strings in a virus name, for every virus name given in the Wikipedia list.

For diseases, we used the list of disease names given in the Unified Medical Language System (UMLS) Metathesaurus. This dictionary has been widely used in disease NER (e.g., Dogan and Lu, 2012; Leaman et al., 2010)\footnote{The list of disease names from UMLS can be found  at https://sites.google.com/site/fmchowdhury2/bioenex .}}.

\item{{\bf Co-Training}: The dictionaries are constructed using the  DL-CoTrain algorithm described previously. The parameters used were $m=5$ and $\epsilon=0.95$  as given in \newcite{collins_co_training}. The phrases present in the \emph{spelling} rules which give a positive label and whose strength is greater than the precision threshold, were added to the dictionary of named entities.

\begin{figure*}	
    \includegraphics[trim=85 530 60 70,clip,width=1.05\textwidth]{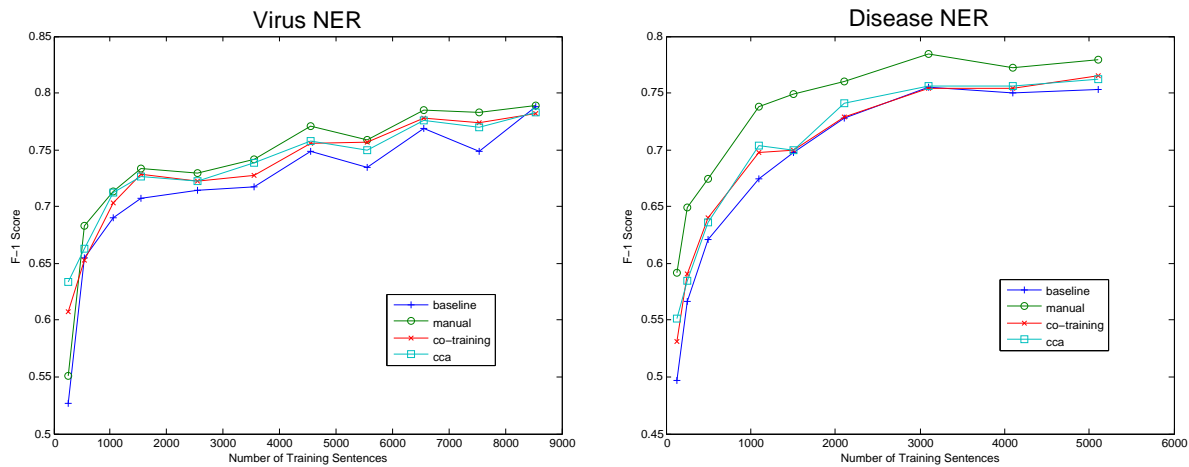}
    \caption{Virus and Disease NER F-1 scores for varying training data size when dictionaries obtained from different methods are injected}	
\end{figure*}

In our experiment to construct a dictionary of virus names, the algorithm stopped after just 12 iterations  and hence the dictionary had only 390 virus names. This was because there were no \emph{spelling} rules with strength greater than $0.95$ to be added. We tried varying both the parameters but in all cases, the algorithm did not progress after a few iterations. We adopted a simple heuristic  to increase the coverage of virus names by using the strength of the \emph{spelling} rules obtained after the $12^{th}$ iteration. All \emph{spelling} rules that give a positive label and which has a strength greater than $\theta$ were added to the decision list of \emph{spelling} rules. The phrases present in these rules are added to the dictionary. We picked the $\theta$ parameter from the set [0.1, 0.2, 0.3, 0.4, 0.5, 0.6, 0.7, 0.8, 0.9] using the development data. 

The co-training algorithm for constructing the dictionary of disease names ran for close to 50 iterations and hence we obtained better coverage for disease names. We still used the same heuristic of adding more named entities using the strength of the rule since it performed better.  }

\item{{\bf CCA}: Using the CCA embeddings of the candidate phrases\footnote{The performance of the dictionaries learned from word embeddings was very poor and we do not report it's performance  here.} as features we learned a binary SVM\footnote{we used LIBSVM (http://www.csie.ntu.edu.tw/~cjlin/libsvm/) in our SVM experiments} to predict whether a candidate phrase is a named entity or not. We considered using 10 to 30 dimensions of candidate phrase embeddings  and the regularizer was picked  from the set [0.0001, 0.001, 0.01, 0.1, 1, 10, 100]. Both the regularizer and the number of dimensions to be used were tuned using the development data.}
\end{itemize}
Table 1 gives the results of the dictionary based taggers using the different methods described above. As expected, when the noisy list of candidate phrases are used as dictionaries the recall of the system is quite high but the precision is very low. The low precision of the Wikipedia virus lists was due to the heuristic used to obtain abbreviations which produced a  few noisy abbreviations but this heuristic was crucial to get a high recall.  The list of disease names from UMLS gives a low recall because the list does not contain many disease abbreviations and  composite disease mentions such as \emph{breast and ovarian cancer}. The presence of ambiguous abbreviations affected the accuracy of this dictionary. 

The virus  dictionary constructed using the CCA embeddings  was very accurate and the false positives were mainly due to ambiguous phrases, for example, in the phrase  \emph{HIV replication}, \emph{HIV} which usually refers to the name of a virus is tagged as a RNA molecule. The accuracy of the disease dictionary produced using CCA embeddings was mainly affected by noisy abbreviations. 
 
We can see that the dictionaries obtained using CCA embeddings perform better than the dictionaries obtained from co-training on both disease and virus NER even after improving the co-training algorithm's coverage using the heuristic described in this section. It is important to note that the dictionaries constructed using the CCA embeddings and a small number of labeled examples performs competitively with dictionaries that are entirely built by domain experts. These results show that by using the CCA based approach we can build NER systems that give reasonable performance even for difficult named entity types with almost no supervision. 

\subsection{Results using a CRF tagger}
We did two sets of experiments using a CRF tagger. In the first experiment,  we add dictionary features to the CRF tagger while in the second experiment we add the embeddings as features to the CRF tagger. The same baseline model is used in both the experiments whose features are described in Section 2.2. For both the CRF\footnote{We used CRFsuite (www.chokkan.org/software/crfsuite/) for our experiments with CRFs.} experiments the regularizers from the set [0.0001, 0.001, 0.01, 0.1, 1.0, 10.0] were considered and it was tuned on the development set.

\subsubsection{Dictionary Features}
\begin{figure*}	
    \includegraphics[trim=85 530 60 70,clip,width=1.05\textwidth]{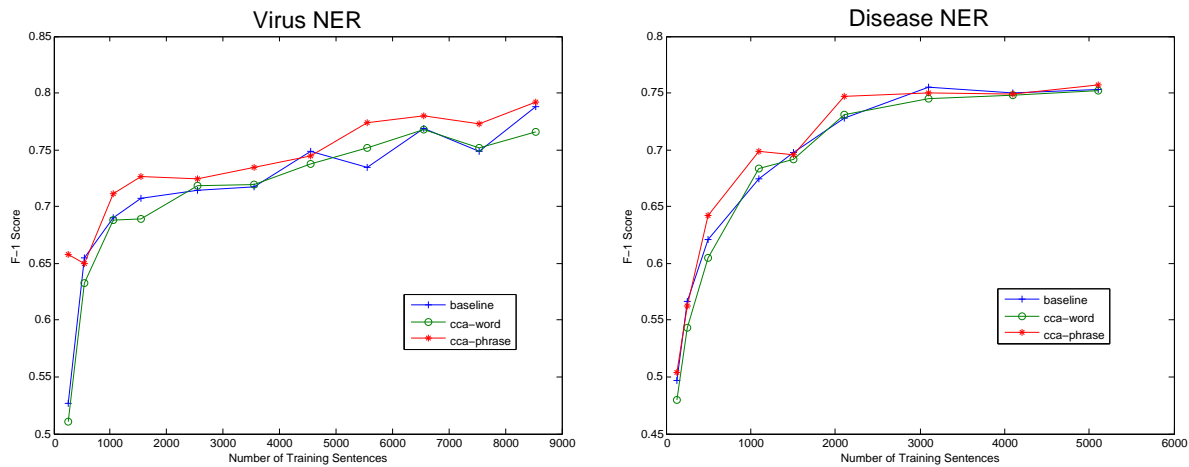}
    \caption{Virus and Disease NER F-1 scores for varying training data size when embeddings obtained from different methods are used as features}	
\end{figure*}

Here, we inject dictionary matches as features (e.g., Ratinov and Roth, 2009; Cohen and Sarawagi, 2004) in the CRF tagger. Given a dictionary of named entities, every word in the input sentence has a dictionary feature associated with it. When there is an exact match between a phrase in the dictionary with the words in the input sentence, the dictionary feature of the first word in the named entity is set to $\texttt{B}$ and the dictionary feature of the remaining words in the named entity is set to $\texttt{I}$. The dictionary feature of all the other words in the input sentence which are not part of any named entity in the dictionary is set to $\texttt{O}$. The effectiveness of the dictionaries constructed from various methods are compared by adding dictionary match features to the CRF tagger. These dictionary match features were added along with the baseline features.
 
Figure 2  indicates that  the dictionary features in general are helpful to the CRF model. We can see that the dictionaries produced from our approach using CCA are much more helpful than the dictionaries produced from co-training especially when there are fewer labeled sentences to train. Similar to the dictionary tagger experiments discussed previously, the dictionaries produced from our approach performs competitively with dictionaries that are entirely built by domain experts.

\subsubsection{Embedding Features}
The quality of the candidate phrase embeddings are compared  with word embeddings by adding the embeddings as features in the CRF tagger. Along with the baseline features, {\bf CCA-word} model adds word embeddings as features while the {\bf CCA-phrase} model adds candidate phrase embeddings as features. {\bf CCA-word} model is similar to the one used in \newcite{dhillon_lrmv}.

We considered adding 10, 20, 30, 40 and 50 dimensional word embeddings as features for every training data size and the best performing model on the development data was picked for the experiments on the test data. For candidate phrase embeddings we used the same number of dimensions that was used for training the SVMs to construct the best dictionary. 

When candidate phrase embeddings are obtained using CCA, we do not have embeddings for words which are not in the list of candidate phrases. Also, a candidate phrase having more than one word has a joint representation, i.e., the phrase ``human immunodeficiency'' has a lower dimensional representation while the words ``human'' and ``immunodeficiency''  do not have their own  lower dimensional representations (assuming they are not part of the candidate list). To overcome this issue, we used a simple technique to differentiate between candidate phrases and the rest of the words. Let $x$ be the highest real valued candidate phrase embedding and the candidate phrase embedding be a $d$ dimensional real valued vector. If a candidate phrase occurs in a sentence, the embeddings of that candidate phrase are added as features to the first word of that candidate phrase. If the candidate phrase has more than one word, the other words in the candidate phrase are given an embedding of dimension $d$ with each dimension having the value $2\times x$. All the other words are given an embedding of dimension $d$ with each dimension having the value $4\times x$. 

Figure 3  shows that almost always the candidate phrase embeddings help the CRF model. It is also interesting to note that sometimes the word-level embeddings have an adverse affect on  the performance of the CRF model. The {\bf CCA-phrase} model performs significantly better than the other two models when there are fewer labeled sentences to train and the separation of the candidate phrases from the other words seems to have  helped the CRF model. 

 \section{Conclusion}
We described an approach for automatic construction of dictionaries for NER using minimal supervision. Compared to the previous approaches, our method is free from overly-stringent assumptions about the data, uses SVD that can be solved exactly and achieves better empirical performance.  Our approach which uses a small number of seed examples performs competitively with dictionaries that are compiled manually. 

\section*{Acknowledgments}
We are grateful to Alexander Rush, Alexandre Passos and the anonymous reviewers for their useful feedback. This work was supported by the Intelligence Advanced Research Projects Activity (IARPA) via Department of Interior National Business Center (DoI/NBC) contract number D11PC20153. The U.S. Government is authorized to reproduce and distribute reprints for Governmental purposes notwithstanding any copyright annotation thereon. The views and conclusions contained herein are those of the authors and should not be interpreted as necessarily representing the official policies or endorsements, either expressed or implied, of IARPA, DoI/NBC, or the U.S. Government.

\bibliographystyle{plain}
\end{document}